\documentclass{sig-alternate}
\newfont{\mycrnotice}{ptmr8t at 7pt}
\newfont{\myconfname}{ptmri8t at 7pt}
\let\crnotice\mycrnotice%
\let\confname\myconfname%

\usepackage[german,american]{babel}
\usepackage[T1]{fontenc}
\usepackage[utf8]{inputenc}
\usepackage{times}
\usepackage{textcomp}
\usepackage{graphicx}
\usepackage{amssymb}
\usepackage{microtype} 

\usepackage{url}
\renewcommand{\tt}{\fontencoding{OT1}\fontfamily{cmtt}\selectfont}

\usepackage[usenames,dvipsnames]{color}
\usepackage{booktabs}
\usepackage[numbers,sectionbib]{natbib}
\makeatletter
\def\@copyrightspace{\relax}
\makeatother

\graphicspath{ {./images/} }

\begin{document}

\title{Using Neural Network for Identifying Clickbaits in Online News Media}
\subtitle{The Albacore Clickbait Detector at the Clickbait Challenge 2017}

\numberofauthors{3}
\author{
\alignauthor
Amin Omidvar\\
\affaddr{Department of Electrical Engineering and Computer Science}\\
\affaddr{York University, Toronto, Canada}\\
\affaddr{omidvar@yorku.ca}\\
\alignauthor
Hui Jiang\\
\affaddr{Department of Electrical Engineering and Computer Science}\\
\affaddr{York University, Toronto, Canada}\\
\affaddr{hj@cse.yorku.ca}\\
\alignauthor
Aijun An\\
\affaddr{Department of Electrical Engineering and Computer Science}\\
\affaddr{York University, Toronto, Canada}\\
\affaddr{ann@cse.yorku.ca}\\
\and
}

\maketitle

\begin{abstract}

Online news media sometimes use misleading headlines to lure users to open the news article. These catchy headlines that attract users but disappointed them at the end, are called Clickbaits. Because of the importance of automatic clickbait detection in online medias, lots of machine learning methods were proposed and employed to find the clickbait headlines. 

In this research, a model using deep learning methods is proposed to find the clickbaits in Clickbait Challenge 2017’s dataset. The proposed model gained the first rank in the Clickbait Challenge 2017 in terms of Mean Squared Error. Also, data analytics and visualization techniques are employed to explore and discover the provided dataset to get more insight from the data. 
\end{abstract}

\section{Introduction}

Todays, headers of news articles are often written in a way to attract attentions from readers. Most of the time, they look far more interesting than the real article in order to entice clicks from the readers or motivate them to subscribe. Online news media publishers rely seriously on the incomes generated from the clicks made by their users, that’s why they often come up with likable headlines to lure the readers to click on the headers. Another reason is that there exists numerous online news media on the web, so they need to compete with each other to gain more clicks from readers or subscription. That’s why most of the online news media have started following this practice. 

These misleading titles, that exaggerate the content of the news articles to create misleading expectations for users, are called clickbaits \cite{anand2017we}. While these clickbaits may motivate the users to open the news articles, most of the time they do not satisfy the expectations of the readers and leave them completely disappointed. Since in the clickbaits, the actual article is of low quality and significantly under-delivers the content promised in the headline, it leads to a frustrating user experience. Moreover, clickbaits damage the publishers’ reputation, as it violates the general codes of ethics of journalism. 

In machine learning and related fields, there have been extensive studies on identifying bad quality content on the web, such as spam and fake web pages. However, Clickbaits are not necessary spam or fake pages. They can be genuine pages delivering low-quality content with exaggerating titles. 

Recently, lots of researches used state of the art machine learning methods to detect clickbaits automatically. Also, some data science competitions for clickbait detection were announced, such as “Clickbait Challenge 2017” \cite{potthast:2017a}, to attract scientists to conduct their researches in this area. In the Clickbait Challenge 2017 competition, different machine learning algorithms were proposed to find the clickbait news headlines. For this particular competition, the goal was to propose a regressor model which can provide a probability of how much clickbait a post is. 

In this research, first the provided datasets are explored and analyzed in order to get more insight from data and to understand the problem better. Then, a deep learning model is proposed which gained 1\textsuperscript{st} ranked in terms of Mean Squared Error on Clickbait Challenge 2017’s dataset \cite{potthast:2017b}. 

\section{Related Work}
In \cite{palau2016reference}, four online sections of the Spanish newspaper El Paris were examined manually in order to find clickbait features that are important to capture readers’ attention. The dataset consists of 151 news articles which were published in June 2015. Some linguistic techniques such as vocabulary and words, direct appeal to the reader, informal language, simple structures were analyzed in order to find their impacts on the attention of the readers.

Two content marketing platforms and millions of headlines were studied to find features that contribute to increasing users’ engagement and change of unsubscribed readers into subscribers. This study suggested that clickbait techniques may increase the users’ engagement temporarily \cite{rony2017diving}.

In \cite{chakraborty2017tabloids}, social sharing patterns of clickbait and non-clickbait tweets to determine the organic reach of the tweets were analyzed. To reach this goal, several tweets from newspapers, which are known to publish a high ratio of clickbait and non-clickbait content, was gathered. Then, the differences between these two groups in terms of customer demographics, follower graph structure, and type of text content were examined.

Natural language processing and machine learning techniques were employed in order to find clickbait headlines. Logistic regression was employed to create supervised clickbait detection system over 10000 headlines \cite{potthast2016clickbait}. They tried to detect clickbait in Twitter using common words occurring in the Tweets through mining of some other tweets’ specific features.

In \cite{anand2017we}, a novel clickbait detection model was proposed using word embeddings and Recurrent Neural Network (RNN). Even though they just considered the headings, their results were satisfactory. Their results gained F1 score of 98\% in classifying online content as clickbaits or not. Furthermore, a browser add-on was developed to inform the readers of diverse media sites regarding the likelihood of being baited via such headlines.

Interesting differences between clickbait and non-clickbait categories which include -but not limited to- sentence structure, word patterns etc. are highlighted in \cite{chakraborty2016stop}. They depend on an amusing set of 14 hand-crafted features to distinguish clickbait headlines. 

Linguistically-infused neural network model was used in \cite{volkova2017separating} to effectively classify twitter posts into trusted versus clickbait categories. They used word embedding and a set of linguistic features in their model. The separation between the trusted and clickbait classes is done by contrasting several trusted accounts to various prejudiced, ironic, or propaganda accounts. At the end, their approach could classify the writing styles of two different kinds of account.  

An interesting model was proposed by Zhou for Clickbait Challenge 2017 \cite{zhou2017clickbait}. He employed automatic approach to find clickbait in the tweet stream. Self-attentive neural network was employed for the first time in this article to examine each tweet’s probability of click baiting. 

Another successful method \cite{grigorev2017identifying}, which was proposed  in Clickbait Challenge 2017, used ensemble of Linear SVM models. They showed that how the clickbait can be detected using a small ensemble of linear models. Since the competitors were allowed to use external data sources, they were used in their research in order to find the pattern of non-clickbait headlines and expand the size of their training set. 

In \cite{glenski2017fishing}, authors developed linguistically-infused network model for the Clickbait Challenge 2017 that is able to learn strength of clickbait content from not only the texts of the tweets but also the passage of the articles and the linked images. They believed using the passage of the articles and the linked images can lead to a substantial boost in the model’s performance. They trained two neural network architectures which are Long Short-Term Memory (LSTM) \cite{hochreiter1997long} and Convolutional Neural Network (CNN). Their text sequence sub-network was constructed using embedding layer and two 1-dimensional convolution layers followed by a max-pooling layer. They initialize their embedding layer with pre-trained Glove embeddings \cite{pennington2014glove} using 200-dimensional embeddings. 

In \cite{thomas2017clickbait}, another model was proposed using neural networks for the Clickbait Challenge 2017. In the text processing phase, they used whitespace tokenizer with lower casing and without using any domain specific processing such as Unicode normalization or any lexical text normalization. Then all the tokens were converted to the word embeddings which were then fed into LSTM units. The embedding vectors were initialized randomly.  They employed batch normalization to normalize inputs to reduce internal covariate shift. Also, the risk of over-fitting was reduced through using dropout between individual neural network layers. At the end, individual networks are fused by concatenating the dense output layers of the individual networks which then were fed into a fully connected neural network. 

A machine learning based clickbait detection system was designed in \cite{cao2017machine}. They extracted six novel features for clickbait detection and they showed in their results that these novel features are the most effective ones for detecting clickbait news headlines. Totally, they extracted 331 features but to prevent overfitting, they just kept 180 features among them. They used all the fields in the dataset such as titles, passages, and key words in their model for extracting these features. 

in \cite{gairola2017neural}, they introduced a novel model using doc2vec \cite{le2014distributed}, recurrent neural networks, attention layers, and image embeddings. Their model utilized a combination of distributed word embeddings and character embeddings using Convolutional Neural Networks. Bi-directional LSTM was employed with an attention layer as well.  

\section{Approach}
\subsection{Data Analytics}

The clickbait challenge’s dataset includes posts from Twitter. Online news media usually use tweeter to publish their links to attract users to their news website. Each post, which is called a tweet, is a short message up to 140 characters that can be accompanied with an image and a hyperlink. Each post is stored in the dataset using JSON object which its structure is described in the table 1.

Human evaluators were employed to assign a clickbait score to each tweet. They had four following options for each tweet:

\begin{itemize}
    \item 
    Score 0: not click baiting (option 1)
    \item
    Score 0.33: slightly click baiting (option 2)
    \item
    Score 0.66: considerably click baiting (option 3)
    \item
    Score 1: clickbait (option 4)
\end{itemize}

\begin{table}
\renewcommand{\arraystretch}{1.3}
\caption{Structure of the JSON object}
\label{tab:example}
\centering
\begin{tabular}{c|p{0.55\linewidth}}
    \hline
    Data Field  &  Description\\
    \hline
    \hline
    ID & The unique ID of the JSON object \\
    \hline
    postText & The text of the tweet\\
    \hline
    postTimestamp & The publish date and time\\
    \hline
    postMedia & The picture that was published with the tweet \\
    \hline
    targetTitle & The title of the linked article \\
    \hline
    targetDescription & Description of the article \\
    \hline
    targetKeywords & The keywords of the actual article \\
    \hline
    targetParagraphs & The content of the actual article \\
    \hline
    targetCaptions & All the captions that exist in the article \\
    \hline
    truthJudgments & Contains 5 scores which were given by human evaluators\\
    \hline
    truthMean & Mean of human evaluators’ scores \\
    \hline
    truthMedian & Median of human evaluators’ scores\\
    \hline
    truthClass & A binary field that indicates the post is clickbait or not 
\end{tabular}
\end{table}

Each tweet was evaluated by 5 evaluators and all the given scores are saved. They provided three datasets for the contesters which one of them does not have labels. Also, they had test dataset for final evaluation of the models which has not been released yet. The information regarding the size of the provided datasets for the participants are shown in the table 2. As we can see in table 2, both datasets 1 and 2 are imbalanced since the number of non-clickbait tweets in datasets 1 and 2 are 2.1 and 3.1 times bigger than the number of clickbait tweets respectively. 

\begin{table}[t]
\renewcommand{\arraystretch}{1.3}
\caption{Statistical Information of the Datasets}
\label{tab:t2}
\centering
\begin{tabular}{c|c|c|c}
    \hline
    Datasets & Tweets & Clickbait & Non-Clickbait\\
    \hline
    \hline
    Dataset 1 & 2495 & 762 & 1697\\
    \hline
    Dataset 2 & 19538 & 4761 & 14777\\
    \hline
    Dataset 3 & 80012 & ? & ? \\
\end{tabular}
\end{table}

So, the target variable that competitors should predict is the mean clickbait score of each post. They did not mention how the binary labels are assigned. It is not based on conventional 0.5 threshold on the mean score since the minimum mean score for the clickbait label is 0.4, and the maximum mean score for non-clickbait label is 0.6. However, the median judgment score is completely in line with the clickbait and non-clickbait labels which is shown in the Figure 1.

\begin{figure}[t]
\label{fg:fg1}
\centering
\includegraphics[width=\columnwidth]{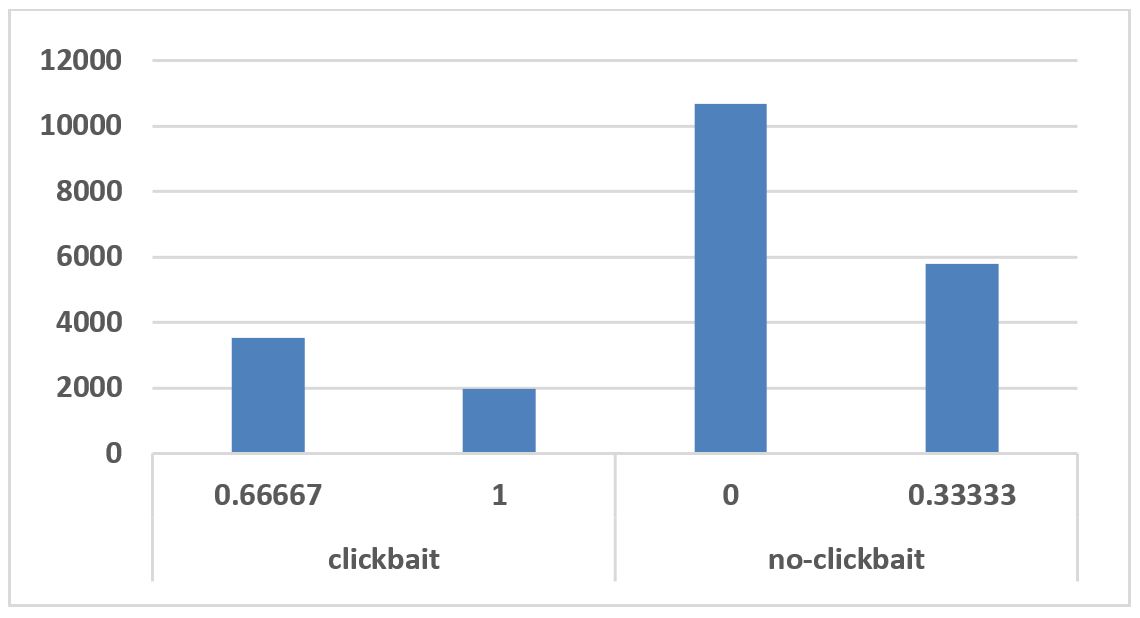}
\caption{Total count of tweets based on the median of the tweets' scores for each binary label}
\end{figure}

As we can see in the Figure 1, all the tweets that their median judgment score is 1 or 0.66667 are in the clickbait category. In contrast, those their median judgment score is 0 or 0.33333 are in the non-clickbait category. So, we can conclude that if the sum of selected “slightly click baiting” and “not click baiting” options is bigger than the sum of two other options, the tweet will be labeled as non-clickbait. Otherwise, it would be considered as a clickbait. 

So, for determining the label of the tweets, there is no difference between option 1 and option 2 (i.e. “not click baiting” and “slightly click baiting”). Also, there is no difference between option 3 and option 4 (i.e. “considerably click baiting” and “click bait”) as well.  

In the Figure 2, min, quartile1, median, quartile3, and max of scores for all the tweets in clickbait and non-clickbait classes are depicted. 

\begin{figure}[b]
\centering
\includegraphics[width=\columnwidth]{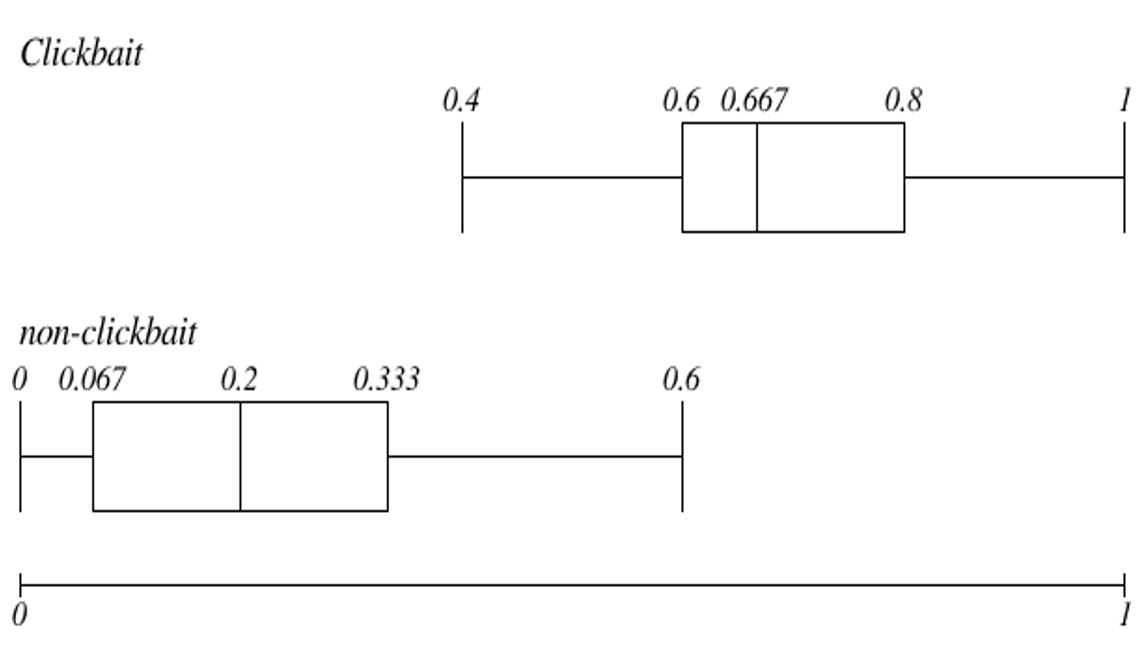}
\caption{Boxplot of the tweets' mean judgement score in each binary class}
\end{figure}

As we can see in the Figure 2, the maximum value for the mean judgement score of the tweets in non-clickbait category is equal to 0.6. Also, there are some tweets in clickbait category which their mean judgement score is below 0.5. 

Figure 3 shows the distribution of mean judgement score for the tweets in both clickbait and non-clickbait categories. It can be seen how clickbait and non-clickbait tweets have overlap between 0.4 and 0.6 values in terms of mean judgement score. 

\begin{figure}[t]
\centering
\includegraphics[width=\columnwidth]{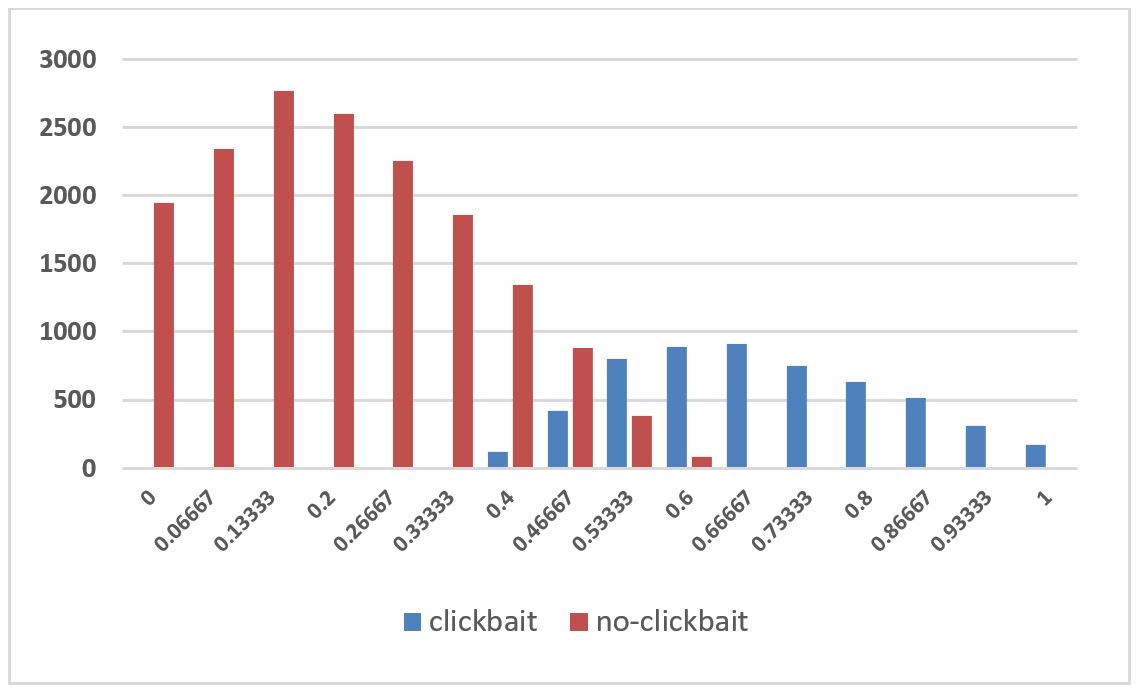}
\caption{Distribution of clickbait and non-clickbait tweets based on mean judgement score}
\end{figure}

Figure 4 shows the distribution of tweets based on their post length with respect to the number of tweets in each class. The vertical axis shows the percentage of tweets in each class while the horizontal one represents the post length. As we can see in the Figure 4, the percentage of short tweets in the clickbait class is higher than the one for the non-clickbait class. 

\begin{figure}[b]
\centering
\includegraphics[width=\columnwidth]{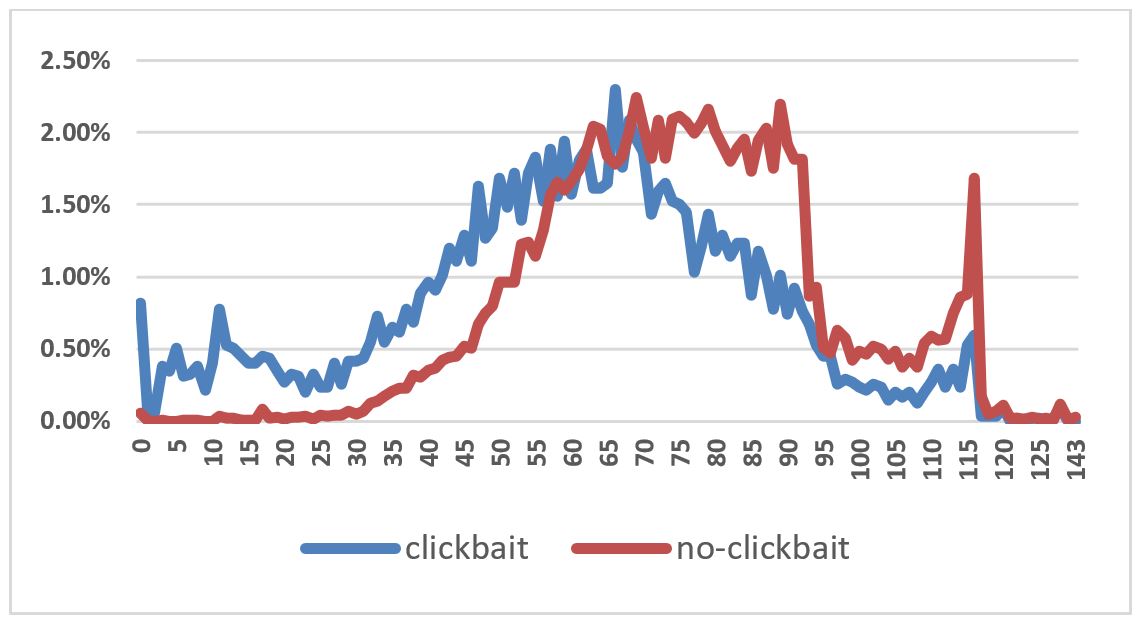}
\caption{The distribution of tweets based on their post length with respect to the number of tweets in each class}
\end{figure}

One of the issue regarding the data set is that while the evaluators are provided with both the post text and a link to the target article, they were not obliged to read the actual article. So, we can consider the judgements are just based on the post texts.  

Also, the scores for the tweets are dependent to the evaluator’s background, knowledge, and topics of interest. So, there should be some noises in the dataset because of the existing differences between evaluators. It was found there are 408 post texts that existed in more than one samples. For example, there are nine samples with the post text “10 things you need to know before the opening bell” which 8 of them were labeled as clickbait and one of them was labeled as non-clickbait. Or there are 14 samples with the post text “Quote of the day:” which two of them are labeled as clickbait.  

\subsection{Proposed Model}

\begin{figure}[t]
\centering
\includegraphics[width=\columnwidth]{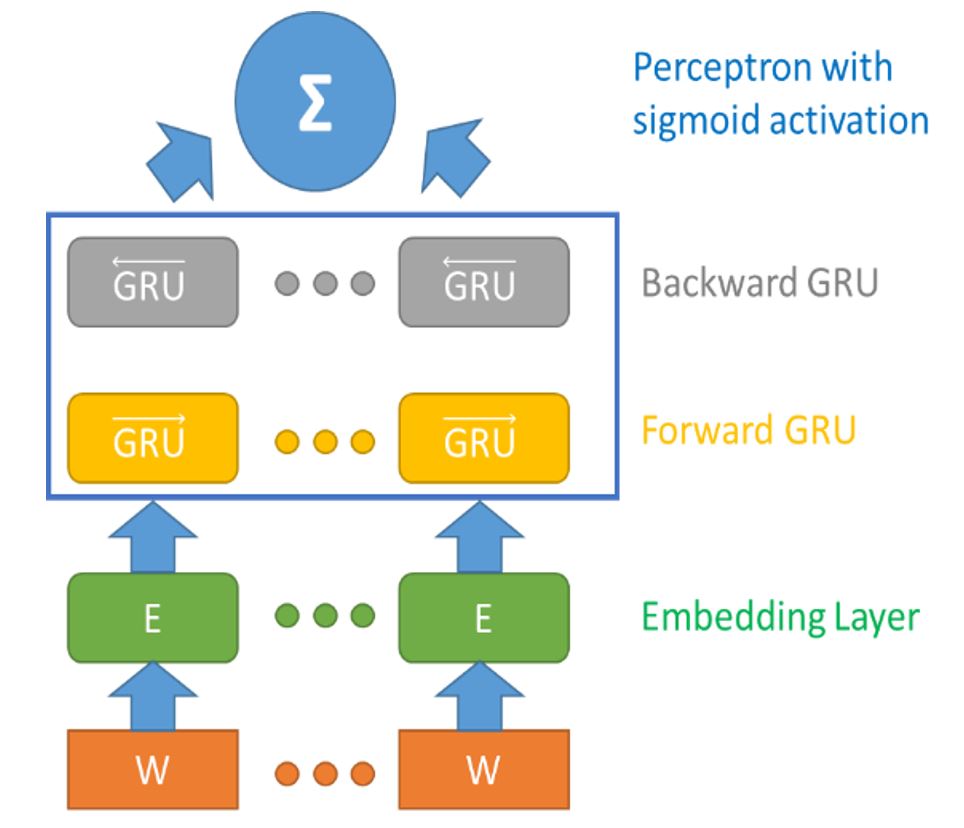}
\caption{The proposed model for clickbait detection}
\end{figure}

To find the best model for clickbait detection, different kind of deep learning architectures were implemented and trained, and their results on our test dataset were compared with each other in order to find the best one among them. For example one of the models was similar to the model that was proposed by Kaveh \cite{taghipour2016neural} for automatic essay scoring which used Convolutional Neural Network along with LSTM to find the scores for each article. The other model was similar to \cite{zhang2015sensitivity} which employed CNNs for text classification task. The model that achieved the lowest Mean Squared Error is shown in the Figure 5.  In this model, we used bi-directional GRU for clickbait detection. Since the test data which was used to compare the contestants’ models has not been published yet, we created our test data set in this research using 30 percent of the Dataset 2 using stratified sampling technique. The models were trained on the training dataset and then they were evaluated using our test data set.  

As we can see in figure 5, the first layer is an embedding layer which transforms one-hot representation of the input words to the their dense representation. We initialized embedding vectors using 50, 100, 200, 300 dimensions using GloVe word embeddings \cite{pennington2014glove}.  

The next layer is a combination of forward GRU and backward GRU. We evaluated bidirectional simple recurrent units \cite{elman1990finding}, bi-directional GRU, and bi-directional LSTM in order to find the best architecture for the clickbait detection task. The result showed that bi-directional GRU outperformed the two other structures.  

The GRU employs a gating approach to trail the input sequences without utilizing separate memory cells. In GRU, there exists two gates which are called update gate z\textsubscript{t} and reset gate r\textsubscript{t}. These two gates are used together in order to handle how to update information for each state. The reset gate and update gate are calculated for each state based on the formula \ref{eq:eq1} and \ref{eq:eq2} respectively. 

\begin{equation}
\label{eq:eq1}
r_t = \sigma (W_r x_t + U_r h_{t-1} + b_r)
\end{equation}

\begin{equation}
\label{eq:eq2}
z_t = \sigma (W_z x_t + U_z h_{t-1} + b_z)
\end{equation}

W\textsubscript{r}, U\textsubscript{r}, b\textsubscript{r}, W\textsubscript{z}, U\textsubscript{z}, b\textsubscript{z} are the parameters of GRU that should be trained during the training phase. The candidate state will be calculated at time t using the formula \ref{eq:eq3}.  

\begin{equation}
\label{eq:eq3}
h_t^\sim = tanh (W_h x_t + r_t \odot (U_h h_{t-1}) + b_h)
\end{equation}

\( \odot \) denotes an elementwise multiplication between the reset gate and the past state. So, it determines which part of the previous state should be forgotten. Finally, formula \ref{eq:eq4} is responsible to calculate the new state.  

\begin{equation}
\label{eq:eq4}
h_t = (1-z_t) \odot h_{t-1} + z_t \odot h_t^\sim
\end{equation}

Update gate in formula \ref{eq:eq4} (i.e. z\textsubscript{t}) determines which information from past should be kept and which new calculated information should be added. The forward way reads the post text from x\textsubscript{1} to x\textsubscript{N} and the backward way reads the post text from x\textsubscript{N} to x\textsubscript{1}. This process is shown through following formulas \ref{eq:eq5} and \ref{eq:eq6}.

\begin{equation}
\label{eq:eq5}
\stackrel{\rightarrow}{h_n} = \stackrel{\rightarrow}{GRU} (x_n, \stackrel{\rightarrow}{h_{n-1}})
\end{equation}

\begin{equation}
\label{eq:eq6}
\stackrel{\leftarrow}{h_n} = \stackrel{\leftarrow}{GRU} (x_n, \stackrel{\leftarrow}{h_{n-1}})
\end{equation}

So, the annotation of the given word x\textsubscript{n} can be calculated through concatenation of the forward state and the backward state which is shown in formula \ref{eq:eq7}

\begin{equation}
\label{eq:eq7}
h_n = [\stackrel{\rightarrow}{h_n},\stackrel{\leftarrow}{h_n}]
\end{equation}

At the end, we used single layer perceptron with sigmoid activation layer in order to figure out the probability of how much clickbait is the tweet. 

\section{Evaluation Results}

We trained our model on “postText”, “targetDescription”, and “targetTitle” separately in order to find which one is better to be used for the clickbait detection task and their results are shown in the Figures 6, 7, and 8 respectively.  
\begin{figure}[b]
\centering
\includegraphics[width=\columnwidth]{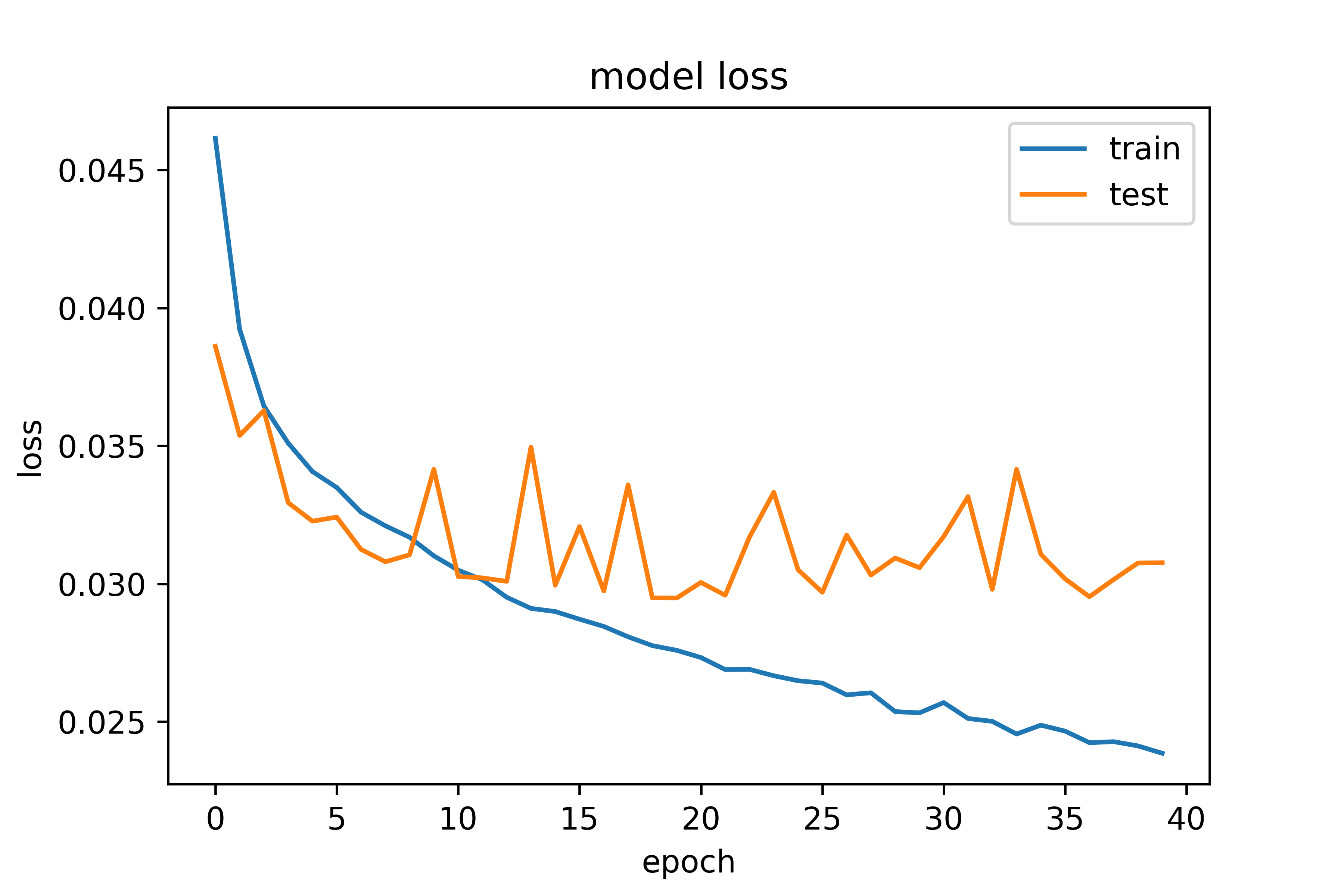}
\caption{Mean Squared Error of the model using “postText” for training}
\end{figure}

\begin{figure}[t]
\centering
\includegraphics[width=\columnwidth]{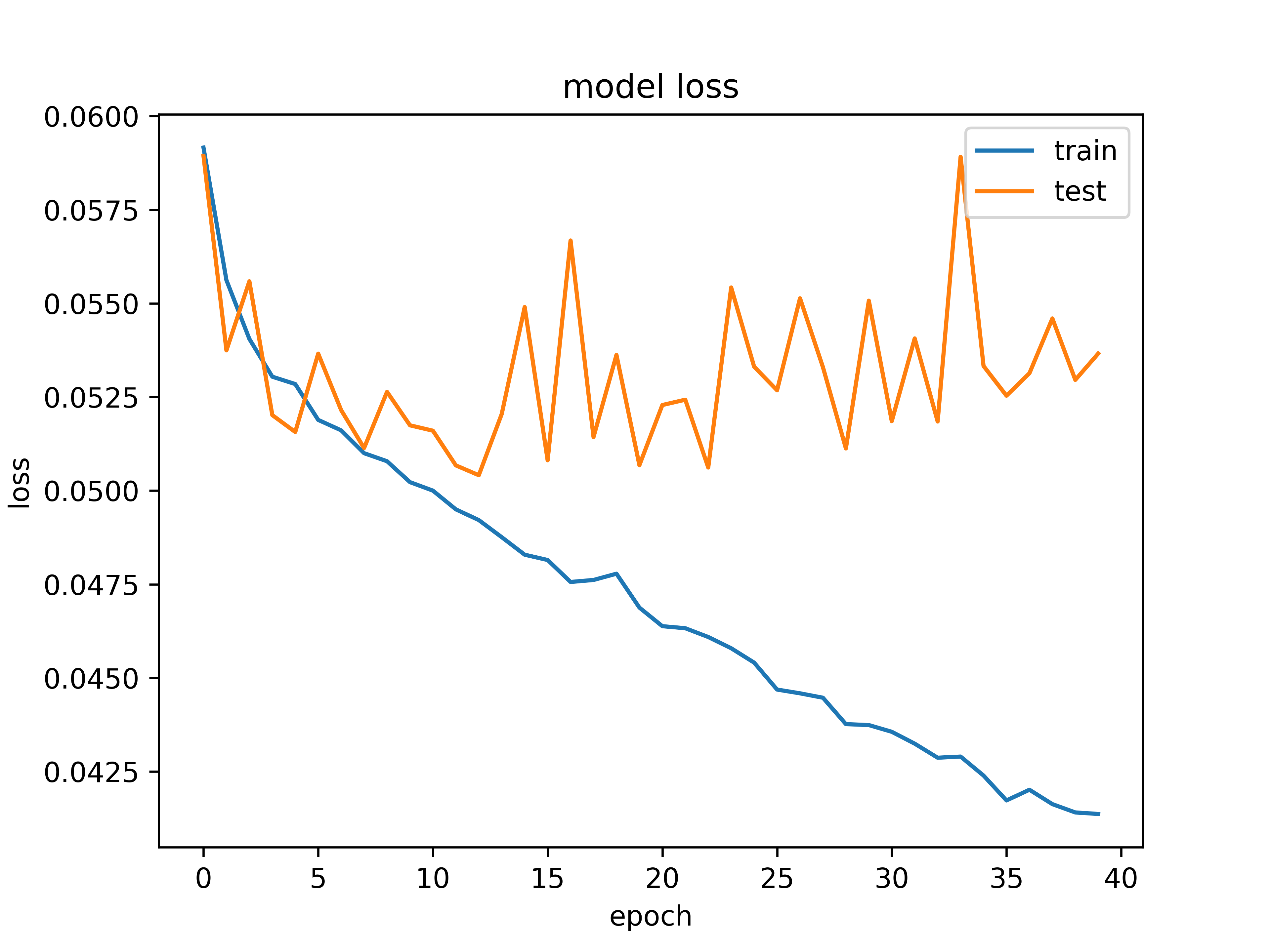}
\caption{Mean Squared Error of the model using "targetDescription" for training}
\end{figure}

\begin{figure}[h]
\centering
\includegraphics[width=\columnwidth]{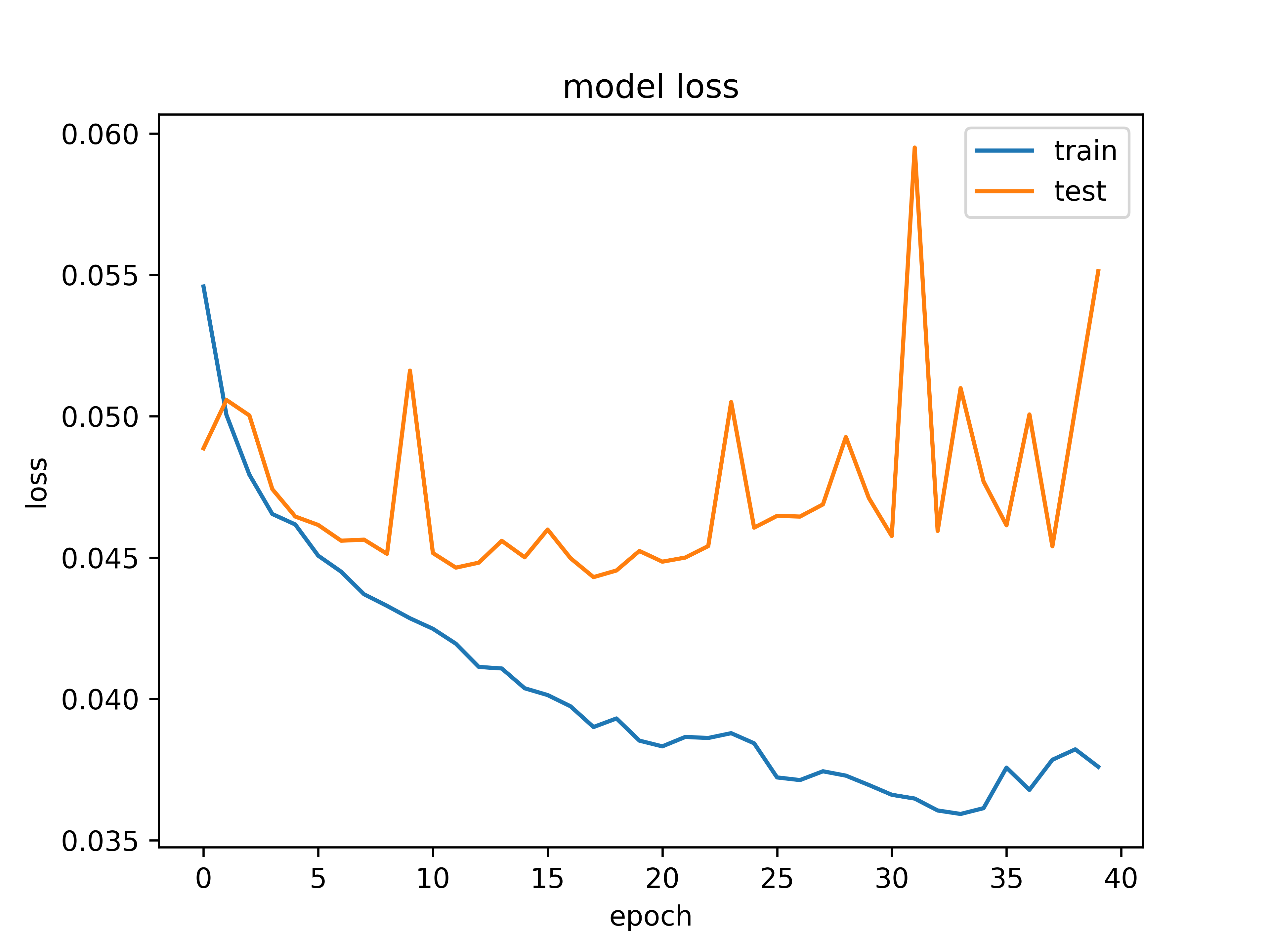}
\caption{Mean Squared Error of the model using "targetTitle" for training}
\end{figure}

As we can see in the Figures 6, 7, and 8, the best result achieved when we trained our model on “postText”. That is because human evaluators did not pay attention to the other data fields for labeling the tweets as much as they paid attention to the “postText”. 

For the training part, mini batched gradient descent with the size of 64 was selected. Mean squared error was selected for the loss function. Drop out technique was employed for Embedding, forward GRU, and backward GRU layers. The model was run with different sets of hyperparameters (i.e. hidden layer sizes, depth, learning rate, embedding vector size, drop out) in order to find the best tuning for the model. Also, the embedding layer was initialized using the Glove embedding vectors with different dimensions. Performance of the models over different number of dimensions was tested, and the result shows 100-dimensions has lower mean squared error in average in comparison with other embedding vectors. 

After tuning the hyper parameters on our own test dataset, we found out the best value for the dropout of the embedding layer is 0.2, and for the input and output of the bi-directional GRU is 0.2, and 0.5 respectively. For the optimizer, we used RMSprob and the size of both forward GRU and backward GRU is 128 which makes the final representation of the tweets a vector with the length of 256.  

Then we used all the available labeled datasets (i.e. test, validation, and training datasets) to train our model. Then the model was run on the Clickbait Challenge’s test dataset using TIRA environment \cite{potthast:2014}. The proposed model gained the first rank among other models in terms of Mean Squared Error. Also, it has the lowest run-time as well. The result of the model on Clickbait Challenge's test dataset is shown in Table 3. 
\\[0.5in]

\begin{table}[t]
\renewcommand{\arraystretch}{1.3}
\caption{Result of the proposed model on Clickbait Challenge's test dataset}
\label{tab:t3}
\centering
\begin{tabular}{c|c}
    \hline
    Mean Squared Error & 0.0315200660647 \\
    \hline
    Median Absolute Error & 0.121994006614 \\
    \hline
    F1 Score & 0.670346813873\\
    \hline
    Precision & 0.731534834992 \\
    \hline
    Recall & 0.618604651163 \\
    \hline
    Accuracy & 0.855261078034 \\
    \hline
    R2 Score & 0.571308479307 \\
    \hline
    Runtime & 00:01:10 \\
    
\end{tabular}
\end{table}

\section{Conclusion}

In this research, the state of the art machine learning algorithms which were proposed for clickbait detection, are introduced. Then a recurrent neural network model was proposed which beat the first ranked model in the clickbait challenge 2017 in terms of the mean squared error measurement. We used mean squared error for model comparison since Clickbait challenge 2017 used this measurement to rank the models.  

The proposed model does not rely on any feature engineering tasks which means they are able to learn the representation automatically in order to classify tweets into clickbait and non-clickbait categories. There exist some very complex models in clickbait challenge 2017 that they did not achieve good result. They tried to utilize all the provided information in the dataset such as images, external linked articles, keywords, etc. to decide whether the headlines are clickbaits or not. In contrast, the proposed model only use “postText” field. Also, the proposed model does not calculate the distribution of the annotations’ probability. Instead of it, just the probability of the clickbait will be calculated which made the proposed model much simpler by converting multi classification task to the binary classification.

\begin{raggedright}
\bibliography{clickbait17-notebook-lit}
\end{raggedright}
\end{document}